\documentclass{article}

\usepackage{babel}
\usepackage[letterpaper,top=2cm,bottom=2cm,left=3cm,right=3cm,marginparwidth=1.75cm]{geometry}

\usepackage{amsmath}
\usepackage{graphicx}

\usepackage{geometry} 
\usepackage{tikz}
\usepackage{alertmessage}
\usepackage{forest}
\usepackage{natbib}
\usepackage[affil-it]{authblk}

\title{On Convolutional Vision Transformers for Yield Prediction}

\author{Alvin Inderka%
  \thanks{Mail: \texttt{s6alinde@uni-bonn.de}}}

\author{Florian Huber%
  \thanks{Mail: \texttt{huber@cs.uni-bonn.de}}}

\author{Volker Steinhage%
  \thanks{Mail: \texttt{steinhage@cs.uni-bonn.de}}}
  
\affil{Department of Computer Science IV, University of Bonn\\ Friedrich-Hirzebruch-Allee 8, 53121, Bonn, Germany}

\begin{document}
\maketitle
\begin{abstract}
While a variety of methods offer good yield prediction on histogrammed remote sensing data, vision Transformers are only sparsely represented in the literature. The Convolution vision Trans\-former (CvT) is being tested to evaluate vision Transformers that are currently achieving state-of-the-art results in many other vision tasks.  CvT combines some of the advantages of convolution with the advantages of dynamic attention and global context fusion of Transformers. It performs worse than widely tested methods such as XGBoost and CNNs, but shows that Transformers have potential to improve yield prediction.
\end{abstract}

\section{Introduction}
Precise yield predictions are used for economic and logistical planning by farmers, traders, producers, and political actors. They can avoid misallocations of food and financial risks for farmers in times of climate change and its impact on yields. While ground-level measurements are cost-intensive and cannot be applied on a large scale, remote sensing data provide a cost-effective and well-studied database of plant condition at regional level. In addition, the \citet{USDA} maintains freely accessible county-level soybean yield data statistics for the United States. Using this data, traditional machine learning methods and deep learning methods could be developed and refined. The often more successful deep learning methods are mainly Artificial Neural Networks (ANN), Deep Neural Networks (DNN), Convolutional Neural Networks (CNN), Long-Short Term Memory Networks (LSTM), CNN-LSTMs and some similar variants of these \citep{muruganantham2022systematic}. The use of more recent vision Transformers for yield prediction, on the other hand, is sparsely represented and will be examined in this brief elaboration using the Convolutional vision Transformer (CvT) by \citet{wu2021cvt}.

\section{Related Work} \label{chap:rel}
A variety of studies addressing yield prediction with remote sensing data and deep learning applications. \citet{muruganantham2022systematic} provides a compact overview of the prevailing deep learning approaches in a systematic literature review. \citet{van2020crop} offers a similar systematic literature review with the additional consideration of traditional machine learning methods. A general overview of Transformers in remote sensing is given by \citet{aleissaee2023transformers}. However, despite the fairly wide scope of the work, regression is not discussed by \citet{aleissaee2023transformers} and, in general, regression by vision Transformers is not well represented in the literature as regression is a rather unusual vision task. 
\subsection{Yield Prediction}
\citet{you} proposed a CNN and a LSTM pipeline for soybean yield prediction in the USA. For the purpose of dimensionality reduction, the satellite data of each county are preprocessed to a normalized histogram for each band and is fed afterwards to the CNN or LSTM. Additionaly \citet{you} proposed an Gaussian Process appended to the CNN and LSTM, which adds tempo-spatial information to the model and improves the error significantly. \citet{sun2019county} combines the advantages of spatial feature extraction of CNNs and the capability of revealing phenological characteristic of LSTMs while using similar histograms. The resulting CNN-LSTM Model extracts the features by a CNN and uses an LSTM on the resulting feature map to learn temporal features. This method proposed by \citet{sun2019county} achieves a higher performance than CNNs and LSTMs individually and enables a satisfactory result at an earlier stage before the harvest. \citet{shooka2020crop} is using a stacked LSTM and a Temporal Attention Model, which basically uses a Temporal Attention Mechanism applied to the stacked LSTM for soybean yield prediction on weather data given by weather.com and yield, genotype and location data given by uniform soybean tests (UST) from the USDA. With the help of a greedy search, the features with the greatest influence on the stacked LSTM are picked out and the weights of the Temporal Attention Mechanism provide information about the temporal dependencies of the Temporal Attention Model, which gives a certain explainability of this deep learning models. The LSTM and the Temporal Attention Model of \citet{shooka2020crop} have a similar performance, but both are more performant than a Random Forest (RF) and LASSO regression. Another approach to promote explainability of yield prediction is an XGBoost approach by \citet{huber2022extreme}, which uses MODIS satellite data and Daymet weather data to propose an explainable model with state-of-the-art performance. \citet{liu2022rice} use a Transformer model named Informer to predict rice yield in the Indo-Gangetic Plains of India. In addition to the official yield data, 8 different sequential variables such as the Normalized Difference Vegetation Index (NDVI) were calculated using MODIS datasets and two other environmental datasets. \citet{liu2022rice} show that the Informer performs better on this task than Attention-based LSTM (AtLSTM), XGBoost, LASSO Regression and a Random Forest (RF), with AtLSTM being the second best method. \citet{bi2023transformer} uses a pipeline of two standard vision Transformers named ViT for feature extraction of near-field images. The images are preprocessed so that one vision Transformer is responsible for soil and one for soybean plants. The respective outputs are multiplied and fed into another Transformer, where they are processed with seed information and fully connected neural networks (FCNN). The resulting performance exceeds the performance of a CNN-LSTM. 

\subsection{Vision Transformer}
As discussed by \citet{aleissaee2023transformers}, there is a variety of vision Transformer models for different remote sensing tasks. Due to this diversity of models and the simultaneous uncertainty regarding the performance of the models in regression and especially yield prediction, only a few widely used models were examined for analysis. Following the success of Transformers in NLP, \cite{dosovitskiy2020image} achieves with Vision Transformer (ViT) an adaptation of the Transformer and its pipeline to images using large training datasets that outperform leading CNNs in some image classification benchmarks. The vision Transformers based on this offer various capabilities and are often presented with state-of-the-art results, especially in the tasks of image classification, object detection and semantic segmenta\-tion.
Based on our previous experience with other methods, two popular vision Transformers in particular were considered for testing. It should be noted that due to the large number of possible vision Transformers and the lack of information on their performance with regard to the yield prediction problem, the selection of the following two Transformers and the choice of the evaluated CvT does not claim to be optimal.
\cite{liu2021swin} present the Swin Transformer, a general purpose Transformer that achieves good performance in image classification, object detection, and semantic segmentation. The Swin Transformer reduces the complexity of ViT, which is quadratic to the image size, to a linear one using a shifted window approach. In this approach, self-attention is only used within non-overlapping windows and the relationships between windows are solved by alternating window partitioning between layers. The Convolutional vision Transformer (CvT) evaluated below originates from \cite{wu2021cvt} and combines the advantages of CNNs and Transformers. CvT is explained in chapter \ref{chap:meth}. \cite{raghu2021vision} analyze the internal representation structure of vision Transformers (using ViT) in comparison to CNNs (using ResNet), showing that the strong inductive bias of spatial equivariance in CNNs is particularly advantageous in the use of local information. ViT can only use this local information effectively with the help of large amounts of data or large-scale pretraining, but has advantages in the global use of information due to self-attention.

\section{Data}
In order to ensure the best possible comparability with other models, the processed dataset from \citet{huber2022extreme} is used and reduced to the test years 2018 to 2021 for an accelerated evaluation. As the data set originates directly from \citet{huber2022extreme}, reference is made to it for a more detailed description. A brief summary of the data and its processing according to \citet{huber2022extreme} is provided below.

The dataset covers the 13 US states with the majority of US soybean acreage. Yield data from the \citet{USDA} is used at district level for the years 2003 to 2021 inclusive. The satellite data comes from NASA's Moderate Resolution Imaging Spectroradiometers (MODIS). In particular, the products MOD09A1 \citep{mod09a1} with 7 bands for spectral surface reflectance and MYD11A2 \citep{myd11a2} with 2 bands for land surface temperature are used. These products contain a resolution of 500 m per pixel and consist of an 8-day composition of the images. The daily average water vapor pressure and the daily absolute precipitation of the Daymet V41 dataset \citep{osti_1148868} are used for the weather. The resolution of 1 km is reduced to 500 m using Google Earth Engines image pyramid and the daily data is aggregated to 8-day compositions. The United States Census Bureau TIGER dataset \citep{Tiger} from 2018 is used to limit the respective satellite data. In addition, an upscaled version of the USDA NASS Cropland Data Layer is used to mask the soybean fields. All satellite data can be found in the Google Earth Engine catalog.

The given data is cropped to an end-of-year harvest period between the 49th and 321st day of the year and assigned to the respective district. The pixels containing soybean fields are then extracted from the satellite data for each district and converted into a histogram per band with 32 equally distributed bins of frequencies. Thus, a resulting data point of a district for one year consists of 11 histograms with a width of 34 8-day intervals and a height of 32 bins. In addition to identifying information on the respective district, the data point is also annotated with the corresponding USDA yield data as ground truth. This results in 14,543 data points with an average yield of 45.26 bushels per acre and a standard deviation of 10.80 bushels per acre.

\section{Methods} \label{chap:meth}
Inspired by the success of CNN models in yield prediction and Vision Transformer in different Tasks, we evaluate the Convolutional vision Transformer (CvT) \citep{wu2021cvt}, which is proposed for image classification. CvT combines the advantages of CNNs (local receptive fields, shared weights, spatial subsampling) and Transformer (dynamic attention, global context fusion, better generalization). 
As we do not make any substantial adjustments to CvT, the following is a summary of the main components of CvT according to \citet{wu2021cvt}. Details can be found in the reference. The adjustments to use CvT as a model for predicting soybean yield using satellite imagery are described afterwards.

CvT as proposed by \citet{wu2021cvt} contains 3 stages with a similar structure. Starting with the input image, in our case our multi dimensional histogram, the Convolutional Token Embedding Layer generates tokens by the use of convolution followed by flattening and layer normalization. These tokens correspond to the tokens, often vectorized words, of the Transformers in NLP. The number and feature dimension of these can be adjusted using the stride of the convolution. The following Convolutional Transformer Block essentially corresponds to the usual procedure of Multi-Head Self-Attention with the modification of the initial linear projections of the input tokens to convolutional projections with depth-wise separable convolutions. To use an convolutional projection, it is necessary to reshape the tokens into a 2 dimensional token map and flatten the result of the convolution again. Three different convolutional projections are used to determine query, key and value. In particular, it is possible to undersample using a stride of 2 for key and value and thus reduce the number of tokens by a factor of 4. Query, key and value are then used for the Multi-Head Attention (MHA) and generate a token map with the help of residual connections, normalization and an MLP, as known from the Multi-Head Self-Attention. The next stage of the same form can be applied to the token and a spatial downsampling can therefore be achieved. A classification token is added to the last stage. The classification token is passed to a final MLP, which finally determines the class or, in our case, the yield.

The adaptation of CvT for use in soybean prediction is basically the adaptation of the input and output of the model and does not require structural changes in this adaption. In addition to the changed input dimension, all augmentation methods must be disabled. The output is changed to one class to represent a regression and its output value is learned using the mean squared error as loss. The evaluation is then carried out using the mean squared error (MSE), its root (RMSE) and the R² metric. The MSE and RMSE should therefore be minimized in their role as error values, while the R² metric should be maximized in its role as a coefficient of determination. All data points from one year (the test year) are used for evaluation and all data points from all previous years are used for training. In addition, to calculate the validation loss and to find the resulting model with the best performance (similar to an early stopping), 10\% of the training data points are randomly separated from the training dataset to form a validation dataset.

\section{Results}
\subsection{Configurations of CvT}
To evaluate the use of CvT for soybean yield prediction, the standardized models according to \citet{wu2021cvt} are tested with the corresponding mostly unchanged hyperparameters. The number of epochs is set to 150 for CvT-13, 200 for CvT-21 and 250 for CvT-W24. After this number of epochs, the loss hardly shows any further changes. The learning rate is manually set to 0.00025, as slightly different learning rates and lower learning rates of the larger models do not result in any advantages.
The evaluation is carried out using the root mean squared error (RMSE) in bushels per acre averaged over the years 2018, 2019, 2020 and 2021. 90\% of the remaining data points in years before the respective test year are used for training and 10\% for validation. To counteract randomness, all results are averaged over 4 runs. The results can be seen in Table \ref{tab:res}. Performance improves slightly as the size of the model increases. At the same time, the runtime increases considerably, meaning that the trade-off between performance and runtime is expensive in terms of runtime.

\begin{table}[ht]
\centering
\begin{tabular}{|c|cc|cc|cc|}
\hline
\textbf{}     & \multicolumn{2}{c|}{\textbf{CvT-13}} & \multicolumn{2}{c|}{\textbf{CvT-21}} & \multicolumn{2}{c|}{\textbf{CvT-W24}} \\ \hline
Year          & RMSE             & R²                & RMSE             & R²                & RMSE              & R²                \\ \hline
\textbf{2018} & 6.64             & 0.580             & 6.34             & 0.618             & 6.38              & 0.613             \\
\textbf{2019} & 5.88             & 0.517             & 5.96             & 0.503             & 5.92              & 0.510             \\
\textbf{2020} & 7.06             & 0.504             & 7.11             & 0.495             & 6.74              & 0.547             \\
\textbf{2021} & 6.61             & 0.747             & 6.14             & 0.783             & 6.35              & 0.767             \\ \hline
\textbf{AVG}  & 6.55             & 0.587             & 6.39             & 0.600             & \textbf{6.35}     & \textbf{0.609}    \\ \hline
\end{tabular}
\caption{Results of the different standard CvT models as RMSE in bushels per acre.}
\label{tab:res}
\end{table}

\begin{table}[hb]
\centering
\begin{tabular}{|c|cc|cc|cc|cc|}
\hline
\textbf{}     & \multicolumn{2}{c|}{\textbf{CvT-13}} & \multicolumn{2}{c|}{\textbf{CvT-13 KV-Stride 1}} & \multicolumn{2}{c|}{\textbf{CvT-21}} & \multicolumn{2}{c|}{\textbf{CvT-21 KV-Stride 1}} \\ \hline
Year          & RMSE             & R²                & RMSE                      & R²                        & RMSE             & R²                & RMSE                      & R²                        \\ \hline
\textbf{2018} & 6.64             & 0.580             & 6.30                      & 0.622                     & 6.34             & 0.618             & 6.67                      & 0.577                     \\
\textbf{2019} & 5.88             & 0.517             & 5.58                      & 0.563                     & 5.96             & 0.503             & 5.73                      & 0.540                     \\
\textbf{2020} & 7.06             & 0.504             & 7.07                      & 0.502                     & 7.11             & 0.495             & 7.00                      & 0.512                     \\
\textbf{2021} & 6.61             & 0.747             & 6.35                      & 0.767                     & 6.14             & 0.783             & 6.22                      & 0.776                     \\ \hline
\textbf{AVG}  & 6.55             & 0.587             & \textbf{6.33}             & \textbf{0.613}            & 6.39             & 0.600             & 6.41                      & 0.601                     \\ \hline
\end{tabular}
\caption{CvT-13 and CvT-21 with a smaller stride of the Convolutional Projection of key and value (KV-Stride; originally it is 2). In this way, advantages in efficiency are cancelled by a larger number of key and value tokens.}
\label{tab:kv_stride}
\end{table}

Due to the use of CvT on histograms with a low resolution of $34\times32$ pixels, the assumption regarding the smoothness of neighboring pixels must be critically examined. This is used to reduce the computational costs in the Convolutional Transformer Block. Key and value are convolutional projected with a stride of 2, so that the following Multi-Head Attention receives a quarter of the originally required tokens for key and value. Table \ref{tab:kv_stride} shows the results of CvT with a stride of 1 of the convolutional projection of key and value and therefore a 4 times more expensive multi-head self-attention. It can be seen that this increases the performance of the small CvT-13 model, which then achieves the best result with a lower runtime than its larger competitors. This pattern does not carry over to CvT-21.

\subsection{Comparison between XGBoost, CNN and CvT}
The CvT-W24 model achieved the best performance of the standard configurations tested. Table \ref{tab:res_comp} compares the RMSE and the R² metric of the CNN according to \cite{you} and the XGBoost method according to \cite{huber2022extreme}. The trivial application of CvT-W24 is subject to both XGBoost and CNN. CvT-W24 has the longest runtime of the compared methods.

\begin{table}[ht]
\centering
\begin{tabular}{|c|cc|cc|cc|}
\hline
\textbf{}     & \multicolumn{2}{c|}{\textbf{XGBoost}} & \multicolumn{2}{c|}{\textbf{CNN}} & \multicolumn{2}{c|}{\textbf{CvT-W24 Plain}} \\ \hline
Year          & RMSE              & R²                & RMSE            & R²              & RMSE                 & R²                   \\ \hline
\textbf{2018} & 4.51              & 0.76              & 6.15            & 0.63            & 6.38                 & 0.613                \\
\textbf{2019} & 4.21              & 0.76              & 5.52            & 0.57            & 5.92                 & 0.510                \\
\textbf{2020} & 4.22              & 0.80              & 6.66            & 0.55            & 6.74                 & 0.547                \\
\textbf{2021} & 4.55              & 0.82              & 5.12            & 0.85            & 6.35                 & 0.767                \\ \hline
\textbf{AVG}  & 4.37              & 0.79              & 5.86            & 0.65            & 6.35                 & 0.609                \\ \hline
\end{tabular}
\caption{Annual results of CvT-W24 in bushels per acre compared to CNN according to \cite{you} (without a gaussian process) and XGBoost according to \cite{huber2022extreme}. The results are taken directly from \cite{huber2022extreme} and take the Daymet data into account.}
\label{tab:res_comp}
\end{table}

\begin{table}[ht]
\centering
\begin{tabular}{|c|cc|cc|cc|}
\hline
\textbf{}     & \multicolumn{2}{c|}{\textbf{XGBoost}} & \multicolumn{2}{c|}{\textbf{CNN}} & \multicolumn{2}{c|}{\textbf{CvT-13}} \\ \hline
Year          & RMSE              & R²                & RMSE            & R²              & RMSE             & R²                \\ \hline
\textbf{2018} & 5.27              & 0.68              & 7.30            & 0.50            & 7.92             & 0.40              \\
\textbf{2019} & 4.65              & 0.70              & 9.73            & -0.49           & 9.75             & -0.33             \\
\textbf{2020} & 5.35              & 0.65              & 8.24            & 0.32            & 8.82             & 0.22              \\
\textbf{2021} & 6.04              & 0.61              & 8.11            & 0.62            & 8.85             & 0.55              \\ \hline
\textbf{AVG}  & 5.33              & 0.66              & 8.35            & 0.24            & 8.83             & 0.21              \\ \hline
\end{tabular}
\caption{Annual results of CvT-13 in bushels per acre compared to CNN according to \cite{you} (without a gaussian process) and XGBoost according to \cite{huber2022extreme} for in-year prediction. The results are taken directly from \cite{huber2022extreme} and take the Daymet data into account.}
\label{tab:res_short}
\end{table}

In order to test possible advantages of CvT in an earlier prediction, an in-year prediction is also carried out. This contains a time span from the 49th to the 201st day of each year and therefore ends before the harvest. The horizontal histogram contains 19 instead of 34 8-day intervals. The results of CvT-13 on this in-year prediction, as well as comparative results from XGBoost and the CNN, can be seen in Table \ref{tab:res_short}. The results show a similar situation to the end-of-year results. 

\section{Conclusion}
The Convolutional vision Transformer shows a lower performance than the competing methods CNN and XGBoost. Even if it were possible to close the gap to the CNN method by a very fine adjustment of the hyperparameters, this would not lead to any advantage over the CNN and the performance would continue to lag behind XGBoost. The higher performance of the CNN compared to the vision Transformer CvT suggests that the local correlations are more important than the global correlations in the histograms. As mentioned in chapter \ref{chap:rel} the use of local features of the CvT could be improved with larger training datasets or large-scale pretraining, which is difficult due to the special input type of low resolution histograms. Other methods such as the SWIN Transformer mentioned in chapter \ref{chap:rel} or transformers that additionally take into account temporal modeling could also improve the performance of the Transformers for yield prediction in further work. Furthermore, it is conceivable that the processing of the comparatively small histograms is unsuitable for exploiting the strengths of the Transformers, which lie particularly in long range dependencies. Therefore, we see a potential for successful application of Transformer in an improved pre-processing of the soybean field pixels for tokenization.

\section*{\begin{normalsize}Acknowledgements\end{normalsize}}
\noindent
This work was partially done within the project ``Artificial Intelligence for innovative Yield Prediction of Grapevine” (KI-iRepro). The project is supported by funds of the Federal Ministry of Food and Agriculture (BMEL) based on a decision of the Parliament of the Federal Republic of Germany. The Federal Office for Agriculture and Food (BLE) provides coordinating support for artificial intelligence (AI) in agriculture as funding organisation, grant number FKZ 28DK128B20. 

\bibliographystyle{plainnat}

\bibliography{references}

\end{document}